\documentclass[preprint,review,12pt]{elsarticle}

\usepackage{amssymb}
\usepackage{amsmath}
\usepackage{array}
\usepackage{booktabs}

\usepackage{graphicx} 
\usepackage{multirow}
\usepackage{fullpage}
\usepackage{tabularx}
\usepackage{rotating, makecell}

\usepackage{algpseudocode}
\usepackage{algorithm}
\usepackage{subfig}

\algnewcommand{\IIf}[1]{\State\algorithmicif\ #1\ \algorithmicthen}
\algnewcommand{\ElseIIf}[1]{\algorithmicelse\ #1} % <<< added
\algnewcommand{\EndIIf}{\unskip\ \algorithmicend\ \algorithmicif}

\newcommand{\abbreviations}[1]{%
  \nonumnote{\textit{Abbreviations:\enspace}#1}}

\usepackage{lineno}

\begin{document}

\begin{frontmatter}
%% Title, authors, and addresses

%% use the tnoteref command within \title for footnotes;
%% use the tnotetext command for theassociated footnote;
%% use the fnref command within \author or \affiliation for footnotes;
%% use the fntext command for theassociated footnote;
%% use the corref command within \author for corresponding author footnotes;
%% use the cortext command for theassociated footnote;
%% use the ead command for the email address,
%% and the form \ead[url] for the home page:
%% \title{Title\tnoteref{label1}}
%% \tnotetext[label1]{}
%% \author{Name\corref{cor1}\fnref{label2}}
%% \ead{email address}
%% \ead[url]{home page}
%% \fntext[label2]{}
%% \cortext[cor1]{}
%% \affiliation{organization={},
%%            addressline={}, 
%%            city={},
%%            postcode={}, 
%%            state={},
%%            country={}}
%% \fntext[label3]{}

\title{Towards RealTime Egocentric Segment Captioning for The Blind and Visually Impaired in RGB-D Theatre Images}

%% use optional labels to link authors explicitly to addresses:
%% \author[label1,label2]{}
%% \affiliation[label1]{organization={},
%%             addressline={},
%%             city={},
%%             postcode={},
%%             state={},
%%             country={}}
%%
%% \affiliation[label2]{organization={},
%%             addressline={},
%%             city={},
%%             postcode={},
%%             state={},
%%             country={}}

\author{Khadidja Delloul\corref{cor1}}
\ead{kdelloul@usthb.dz}
\author{Slimane Larabi}

\affiliation{organization={RIIMA Laboratory, Computer Science Faculty, USTHB University},
            addressline={BP 32}, 
            city={El Alia},
            postcode={16111}, 
            state={Algiers},
            country={Algeria}}
\cortext[cor1]{Corresponding author.}
\abbreviations{Red, Green, Blue, Depth: RGB-D; Feature Pyramid Network, FPN; Region Proposals Network: RPN}

\begin{abstract}
%% Text of abstract
In recent years, image captioning and segmentation have emerged as crucial tasks in computer vision, with applications ranging from autonomous driving to content analysis.
Although multiple solutions have emerged to help blind and visually impaired people move around their environment, few are applications that help them understand and rebuild a scene in their minds through text. Most built models focus on helping users move and avoid obstacles, restricting the number of environments blind and visually impaired people can be in.

In this paper, we will propose an approach that helps them understand their surroundings using image captioning. The particularity of our research is that we offer them descriptions with positions of regions and objects regarding them (left, right, front), as well as positional relationships between regions,  while we aim to give them access to theatre plays by applying the solution to our TS-RGBD dataset.

\end{abstract}
\begin{keyword}
Image Captioning \sep Egocentric Scene Description \sep Dense Captioning \sep Segmentation \sep RGB-D images
%% keywords here, in the form: keyword \sep keyword

%% PACS codes here, in the form: \PACS code \sep code

%% MSC codes here, in the form: \MSC code \sep code
%% or \MSC[2008] code \sep code (2000 is the default)

\end{keyword}

%%Research highlights
%\begin{highlights}
%\item A novel aspect of image captioning defined as egocentric segment captioning.
%\item Each segment of the image has a sentence that describes it. 
%\item The output captions are enriched with positional relationships between regions.
% \item Positional relationships are determined by an algorithm, making the process independent from any machine or deep learning modules. The algorithm is faster and less complex.
%\item A novel application domain for image captioning: RGB-D theatre images.
% \item The proposed solution demonstrates real-time speed during testing.
%\end{highlights}

\end{frontmatter}

%\linenumbers

%% main text
\section{Introduction}
\label{sec:intro}
With the remarkable advancements in deep learning technologies, numerous applications have emerged to assist blind and visually impaired individuals in various aspects of their lives. These range from tools designed to aid in navigation \cite{zatout1, zatout2} and obstacle detection to applications that identify currency bills, and objects, and provide reading assistance or online support.

While these applications offer valuable assistance in daily life transactions and challenges, they primarily focus on the physical aspect of assisting visually impaired individuals in navigating their environment. As a result, the spatial freedom for a blind person is restricted to either indoor or outdoor scenes. Furthermore, these solutions primarily provide information about the presence of an obstacle without conveying detailed information about its nature or appearance.

In addition, they have limitations when it comes to entertainment. Specifically, there is a lack of solutions that enable blind and visually impaired individuals to access and understand theater scenes by providing descriptions of the scene and actors' actions on stage. While projects exist that focus on describing paintings and aesthetics or reading books, to our knowledge, there is an absence of research interested in providing textual descriptions of theater plays. Although manual textual descriptions read by people are sometimes available, they are not always accessible.

Our aim is to build an intelligent system that provides users with visual impairments with a detailed audio description of a given scene. To do that, we had to build a system that combines multiple computer vision fields namely image captioning and image segmentation. 

In this paper, we present a novel approach to image captioning that introduces the concept of segment or region captioning. Our proposed method generates captions that provide egocentric information, offering the directional details of each region within the image. Additionally, our captions combine textual descriptions with both egocentric positions and positional relationships, providing a comprehensive understanding of the image content.

To enable the detection of positional relationships between objects, we propose an algorithm that operates independently of deep learning solutions. This algorithm effectively identifies the spatial connections between different regions, enhancing the richness and contextual understanding of the generated captions.

To evaluate the effectiveness of our solution, we apply it to RGB-D images from our newly developed dataset, TS-RGBD \cite{datasetpaper}. This dataset incorporates RGB images along with depth maps collected using the Microsoft Kinect v1 sensor. By utilising these multi-modal inputs, we introduce a novel field of application for image captioning: Theatre Scenes.

Motivated by the success of \textit{MindsEye Radio} broadcasters \cite{mindseye} in translating visual events to audio for individuals with visual disabilities, our research builds upon this concept and extends it to the domain of image captioning. Our goal aims to enable visually impaired individuals to gain a more immersive experience in theater environments by automatically providing descriptive captions with angle guidance and highlighting the positional relationships between regions. Our work will empower visually impaired individuals with enhanced accessibility and a more engaging experience in appreciating theatre plays.

Generated captions are read to the users using a module that transforms text into audio.

To sum it up, in this work, we make the following contributions: 
\begin{itemize}
    \item We propose a novel aspect of image captioning defined as segment or region captioning;
    \item Generated captions are egocentric, meaning they give the directional information of each region;
    \item Generated captions combine textual descriptions with egocentric positions, as well as positional relationships;
    \item We propose an algorithm to detect positional relationships between objects independently from deep learning solutions;
    \item Our solution is applied to RGB-D images from our TS-RGBD where we use RGB images as well as depth maps that we collected using a Microsoft Kinect v1 sensor, and thus we propose a novel captioning field of application: Theatre Scenes.
\end{itemize}

Through this study, we contribute to the advancement of image captioning techniques, expanding their applicability to specific domains and addressing the needs of diverse user groups.

% \section{Literature Background}
% \label{sec:litbgd}
% \input{sections/litterature_background}

\section{Problematic}
\label{sec:prblm}
This work draws inspiration from \textit{MindsEye Radio} \cite{mindseye}, a platform that offers audio translations of visual events, videos, and news throughout the day. It serves as a valuable resource for individuals with various visual disabilities, providing timely access to information. 

Our research focuses on developing an intelligent system capable of automatically translating the content of an image into descriptive text, a task known as image captioning. However, our system goes beyond generating a single sentence or paragraph; it aims to provide descriptions for each region present in the image, including background, non-salient and salient objects. 

These textual descriptions are designed to be egocentric, providing users with information about the position of each region relative to themselves (e.g., on their left, right, or in front of them), as well as the positional relationships between objects. Our solution would incorporate the determination of the angle to which the user should look to face each region. By providing this angle information, visually impaired individuals can have a more immersive experience as they navigate and explore the theatrical environment.

This angle guidance as well as the positional relationships between regions enhances their understanding of the spatial layout and helps them orient themselves towards the regions of interest within the theatre setting.

Inspired by the translation of sports and art events by \textit{MindsEye Radio} \cite{mindseye}, our solution specifically applies to our TS-RGBD Theatre Images \cite{datasetpaper}, aiming to offer visually impaired individuals the opportunity to experience and appreciate theatre plays.

\section{Related Works}
\label{sec:relworks}
From section \ref{sec:prblm} we gather that our problem bridges three fields of computer vision: \textit{Image Captioning, Image Segmentation,} and \textit{Objects Spatial Relationships}.

In the following, we will go over recent and relevant studies from each one of them.

\subsection{Image Captioning}
Image captioning serves as a bridge between computer vision and natural language processing, wherein an intelligent system takes an image as input and generates a corresponding text description that accurately portrays the content depicted in the image.

Image captioning models can output one sentence for a given image, a paragraph, or multiple captions of each region of interest.

\subsubsection{Sentence Captioning}
Single-sentence captioning models can be classified according to their architectures; image analyzing models, attention mechanisms and transformers, CNN-LSTM networks, or even GANs \cite{mineisia}.

These solutions, whether they rely on transformers and attention mechanisms \cite{ref4, ref9, ref16}, or scene graphs as presented in \cite{ref33}, in which learning is supervised, or relying on beam search analysis or gated recurrent units (GRU) units, in which learning is unsupervised \cite{ref32, ref6}, generate one single sentence for each input image. Such models are trained on RGB image datasets \cite{ref27, ref29}.

\subsubsection{Paragraph Captioning}
One sentence is not enough to describe an image's semantic and contextual content, which motivated researchers to improve single-sentence image captioning to output a describing paragraph for an input image.

Solutions for paragraph captioning are based on single-sentence captioning to generate a set of sentences that will be combined to form a coherent paragraph \cite{mineisia}. Such models are built upon recurrent network or encoder-decoder architectures  \cite{ic20, ic21, ref3, ref11}. 

Some proposed methods \cite{ref14} output paragraphs containing positional relationships between objects, but these relationships are vague and sentences lack details about the described region of interest of the image.

\subsubsection{Dense Captioning}
Dense captioning is the task of textually describing multiple detected regions of interest within an input image by generally using an RPN.

One of the most relevant works is the DenseCap model proposed in \cite{densecap}. It uses an RPN to detect $k$ regions of interest within an image, then each region will be given a caption by the LSTM language network. 

The DenseCap model was trained on RGB images of the Visual Genome dataset, a set of MS-COCO and Flickr indoor or outdoor images.

\subsection{Regions Positional Relationships}
While image captioning with its different output forms provides users with textual descriptions of the content of an image, these descriptions do not always give information about spatial or positional relationships between objects or regions, which became a topic for enriching image captions.

Multiple studies rely on RGB images to describe such relations \cite{ref1, ref7}, nevertheless, without the depth information and without the point cloud, such solutions are non-deterministic and must rely on complex deep learning architectures.

Most relevant works to answer the problem of spatial relationships between regions are being applied to RGB-D datasets. However, models like the ones proposed in \cite{ref15, ref20} use custom datasets or datasets of RGB-D scans, meaning that they were trained only on indoor scenes, with heavy input data and complex architectures.

Models that are trained on public RGB-D datasets tend to focus only on retrieving the relationships between objects with no details on their shapes or appearances \cite{ref19} when generating descriptive paragraphs.

Kong et al proposed in \cite{kong} a solution for RGB-D image captioning, but it only focuses on enriching descriptions by positional relationships between objects, while training their model on a dataset that does not include theatre images. 

\section{State-of-the-Art Limitations}
\label{sec:lim}
Having thoroughly examined the related works in the preceding section, we have identified noteworthy gaps and limitations within the cited literature. In this section, we will shed light on the primary limitations of the existing research, providing critical insights that pave the way for the novel contributions and advancements put forth in our study.

\subsection{Regarding Image Captions}
\subsubsection{Sentence Captions}
Models that generate one sentence to describe an input image tend to focus on mobile objects, and salient regions, and completely ignore any background information and static objects, which makes the sentence less informative about what is present in the scene. Background description is important for a visually disabled person to understand their environment.
\subsubsection{Paragraph Captions}
While paragraph captions provide more information about an input image than sentence captions, paragraph generation models often focus on objects on movement, and rely on background to only name the environment. Such models generally do not include descriptions of positional relationships between objects.

\subsubsection{Dense Captions}
While dense captioning provides detailed captions for each region of interest in an input image, it introduces challenges when it comes to the practical use of the framework. Specifically, when generating captions for multiple regions $k$ in an image, the presence of overlapping bounding boxes can lead to redundancy and duplication of information.

Visually impaired users, who are the intended users of our research, would face limitations in providing $k$ as an input to eliminate unwanted regions and choosing the wanted ones from dense captioning results, due to their lack of awareness regarding the content of the scene.

\subsection{Regarding Datasets}
Our solution focuses on applying image captioning techniques to RGB-D Theatre images. Although these images are not available in mainstream computer vision datasets that typically include generic images, we aim to adapt and tailor the image captioning approach to cater specifically to the unique context of theatre plays.

Our solution will be applied to our newly collected and annotated dataset “TS-RGBD”: a dataset of RGB-D images for Theatre Scenes \cite{datasetpaper}.

\subsection{Regarding Positional and Spatial Relationships Extraction}
As explained in section \ref{sec:relworks}, models that are built to extract positional relationships between objects are trained on either private RGB-D scans of specific indoor scenes or even 3D modeled scenes, or public datasets with only outdoor scenes, thus the absence of theatre images in said datasets.

Such models focus only on retrieving the relationship while completely ignoring the appearances and shapes of objects, which does not solve the problem in our case as we need to output textual descriptions of regions of interest.

Based on these limitations and our inspiration, we introduce the following solution.

\section{Proposed Solution}
\label{sec:propsol}
Our solution consists of four main steps: \textit{i) Panoptic Segmentation, ii)  Segment Captioning, iii) Determining  Region Directions, iv) Determining Positional Relationships.}

\begin{figure*}[!ht] 
    \centering
    \includegraphics[width=15cm]{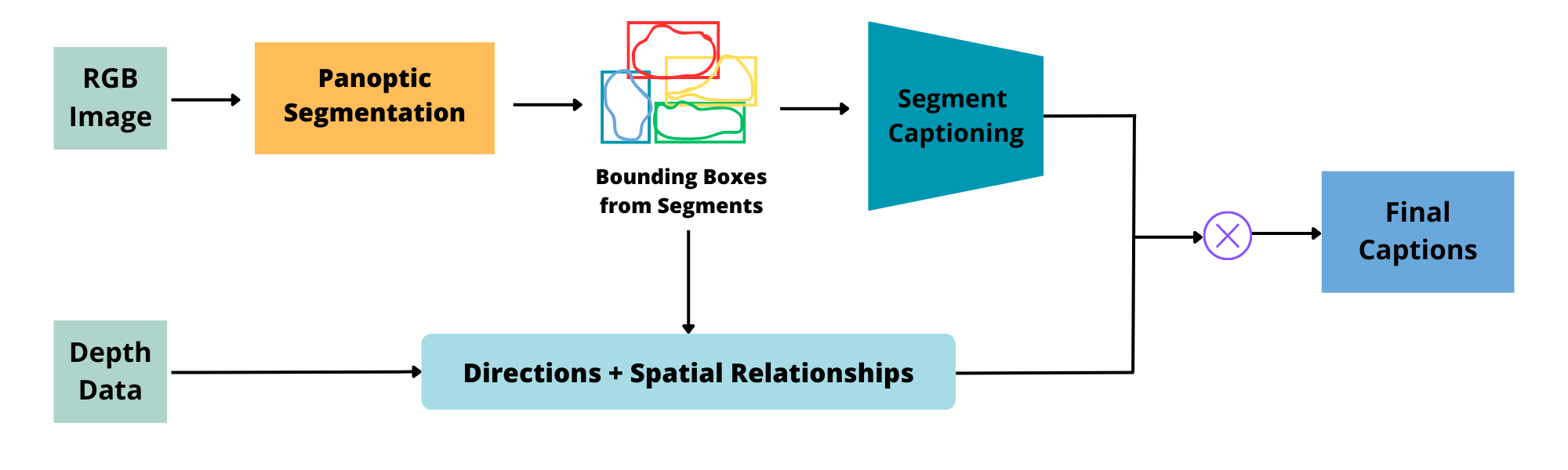}
    \caption{General solution architecture.}
    \label{fig:framework}
\end{figure*}

Figure \ref{fig:framework} illustrates the general architecture of our framework.

The overall architecture of our solution can be described as follows:

\begin{itemize}
    \item The RGB image serves as the input to the panoptic segmentation module. This module analyses the image and generates segmentation results, from which we would get bounding boxes that delineate different regions within the image;
    \item The segment captioning module takes the generated bounding boxes as input, as well as the image, and generates descriptive captions for each region. These captions provide detailed descriptions of the identified segments;
    \item Additionally, our solution incorporates directions and positions algorithms that utilise the depth map and the boxes as input. These algorithms process the depth information from each box to determine spatial relationships between segments;
    \item  The output captions from the segment captioning module are combined with the results from the directions and positions algorithms. This combination results in comprehensive captions that encompass both the detailed segment descriptions and the contextual information derived from the depth map.
\end{itemize}

In summary, our solution uses the RGB image to perform panoptic segmentation, generates bounding boxes for each segment then generates descriptive captions, and combines them with information derived from depth maps to produce complete and informative captions.

In the following, we will detail each step of our solution.

\subsection{Segmentation}
Our goal is to first generate a descriptive caption for each region present in the image. 

Instance segmentation provides segments identifying and delineating individual objects within an image with precise boundaries, so it eliminates background and is only interested in salient objects.

Semantic Segmentation labels each pixel in an image with its corresponding class, and so multiple objects or regions belong to the same segment.

Both instance segmentation and semantic segmentation are not suitable solutions for our problem. Therefore, we have opted for panoptic segmentation, where each pixel is assigned a specific identifier unique to each instance, and a code corresponding to its class.
With panoptic segmentation, we have as output every region on the image, be it salient, background, or non-salient objects.

Regions are not divided into a set of objects but represented as a whole by segments.

\subsection{Segment Captioning}
To generate textual descriptions for each segment, we needed to extract segment features that would feed the language network. 
We developed a solution with two different heads.

\subsubsection{Segment Captioning Head 1}
Initially, we proposed a model that would take as input an image with bounding boxes of  ground truth segments, then it would put all pixels in the box that do not belong to the segment to 0 to cancel them when training the feature extractor. 

\begin{figure*}[!ht] 
    \centering
    \includegraphics[width=15cm]{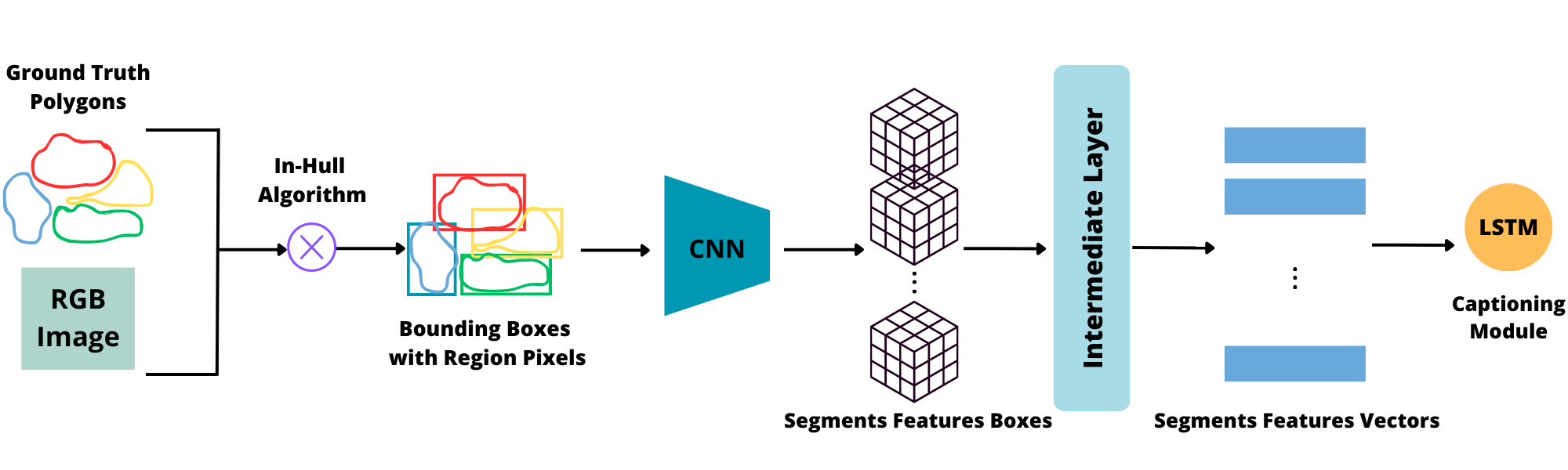}
    \caption{Illustration for the architecture of the first head.}
    \label{fig:head1}
\end{figure*}

Figure \ref{fig:head1} presents the architecture of the proposed solution with the first head.

To check if a pixel belongs or not to a polygonal envelope, we used an optimization algorithm that solves a linear problem to check if a point can be expressed as a convex combination of another set of points. Each pixel from the bounding box passes through the 'in\_hull' algorithm, if it belongs in the polygonal envelope it would be kept, if not, it would be put to 0. Pixels put to 0 are canceled in the CNN features extractor.

Each box would be passed through VGG-16, to extract a map for features.

\subsubsection{Segment Captioning Head 2}
Afterwards, we proposed a modified DenseCap architecture. Densecap consists of three parts \cite{densecap}:
\begin{enumerate}
    \item \textit{The Feature Extractor:} a pre-trained model for features extraction (VGG-16, Resnet-50, FPN with Resnet-50...etc), which extracts features map from an input image;
    \item \textit{RPN:} a region proposal network that would detect multiple regions of interest per image, it returns bounding boxes and scores for each, hence the word "Dense";
    \item \textit{RoI Align:} "Region of Interest Alignment" to get features of each bounding box detected by the RPN from the features map of the input image;
    \item \textit{Intermediate Layer:} that transforms each region features extracted by the RoI Align into a vector of a dimension $D$;
    \item \textit{Language Network (LSTM):} that takes features vectors as input and generates a sentence for each.
\end{enumerate} 

The second part is considered to be a localisation layer for the DenseCap model. We modified it by removing this layer and giving the bounding boxes of segments as input.

\begin{figure*}[!ht] 
    \centering
    \includegraphics[width=15cm]{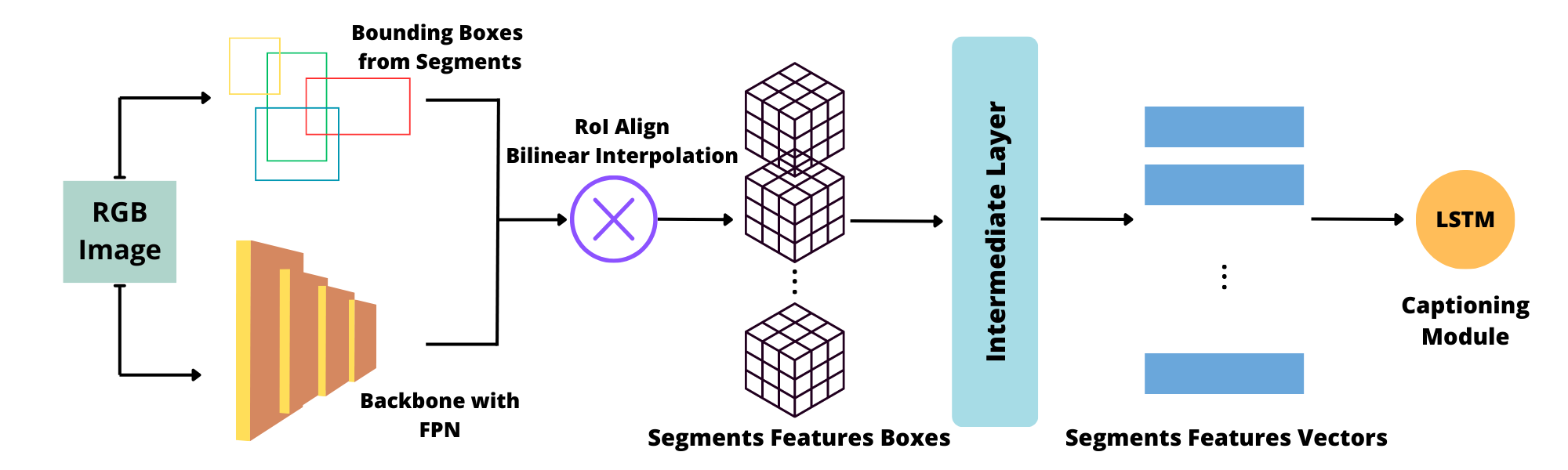}
    \caption{Illustration for the architecture of the second head: a modified version of DenseCap.}
    \label{fig:head2}
\end{figure*}

Figure \ref{fig:head2} presents the architecture of the proposed solution with the second head.

Features were extracted using a pertained Feature Pyramid Network with Resnet-50 backbone. It generates 4 maps with 256 features. The output would have the size $4\times256$.

\subsection{RoI Multi-Scale Alignment}
In order to retrieve the bounding boxes from the features maps, we used the multi-scale region of interest alignment that relies on bi-linear interpolation to extract each bounding box features given its coordinates on the input image. 

Each segment bounding box of size $X \times Y$ is projected onto the 4 grids of convolutional features, and bi-linear interpolation from \cite{bilinear} is used to interpolate and get features of size $X$' $\times Y$' for each segment.

\subsection{Intermediate Layer}
The intermediate layer is a fully connected layer that transforms $B \times C \times size \times size $ features into a vector of $D$ features, where $B$ is the number of boxes, $C$ is the number of extracted features, and $D$ is the length of the output vector.

The features from each region are flattened into a vector and passed through two full-connected layers, each using rectified linear units and regularized using Dropout.

It helps to represent each region as a compact vector of features, instead of a multidimensional matrix.

\subsection{LSTM Network for Captioning}
The DenseCap RNN model for captioning was used, with one  recurrent layer that takes as input the $B \times D$ vectors and outputs $B$ sentences, where $B$ is the number of boxes per image.

The input of this model is a sequence of feature vectors from the intermediary layer, the first token is the START token, and the last is the END token, encoded as $<$bos$>$ and $<$eos$>$ respectively. Each token from the vocabulary is given a code.

The LSTM network computes a sequence of hidden states and outputs vectors using a recurrence formula \cite{densecap}. Output vectors constitute a sequence of tokens for each generated sentence. 

\subsection{Determining Region Directions}
Our solution includes determining the direction for visually impaired individuals to face each described region in the RGB-D Theatre images.

The depth information is important because if we rely only on the 2D image, $(x, y)$ coordinates represent the projections of every point from the real world to the image plane, which can mislead the directional algorithm. So it is best to transform those coordinates into a 3D point cloud using the depth map. 

\begin{figure}[!ht] 
    \centering
    \includegraphics[width=7cm]{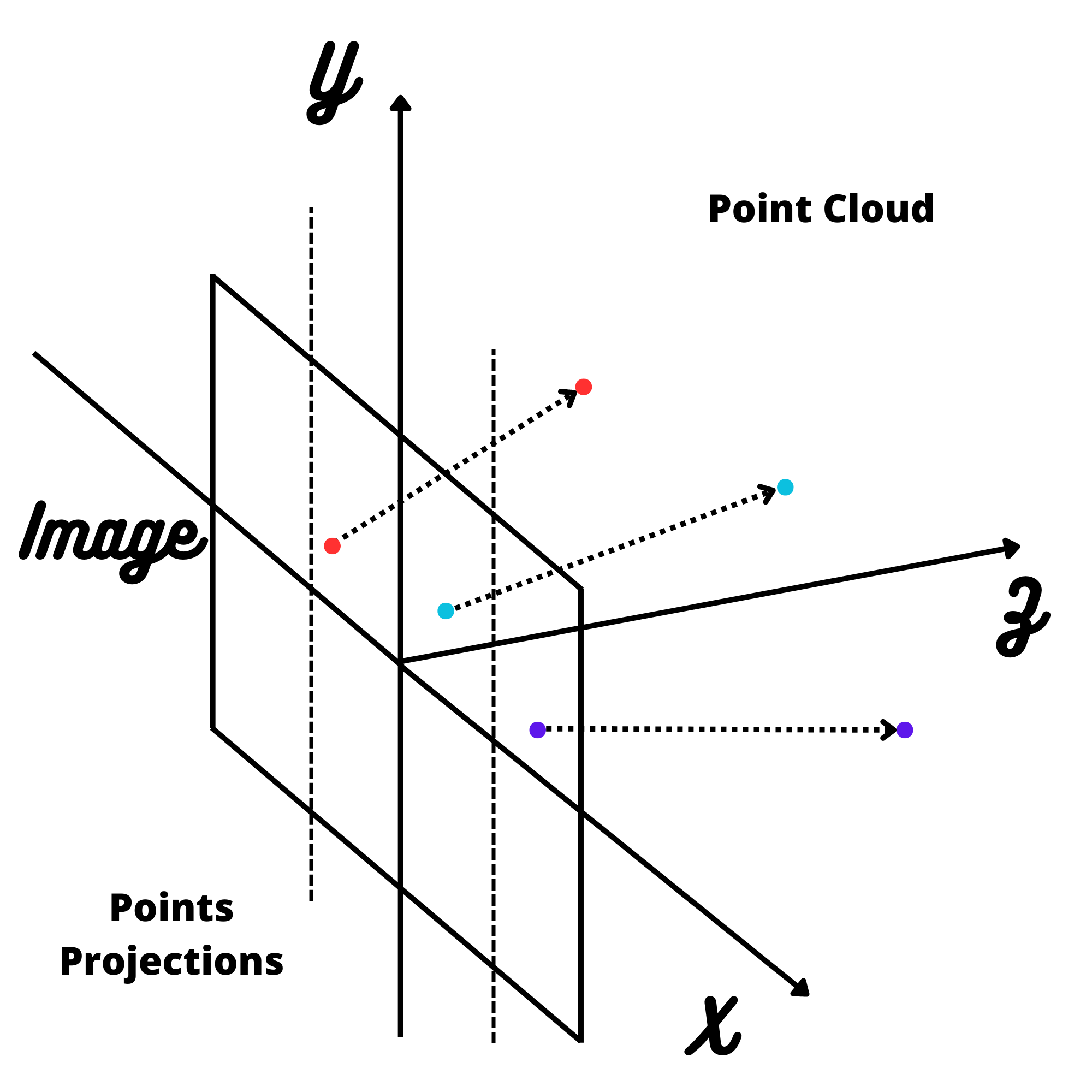}
    \caption{Projections of points from the real world to the image change the positions of pixels.}
    \label{fig:projections}
\end{figure}

Figure \ref{fig:projections} shows how the red point may be considered upfront when looking only at the image, but it is in reality on the left.

So to retrieve the direction of a region we need to get the angle that its centroid forms with the $(X)$ axis.

\begin{figure}[!ht] 
    \centering
    \includegraphics[width=7cm]{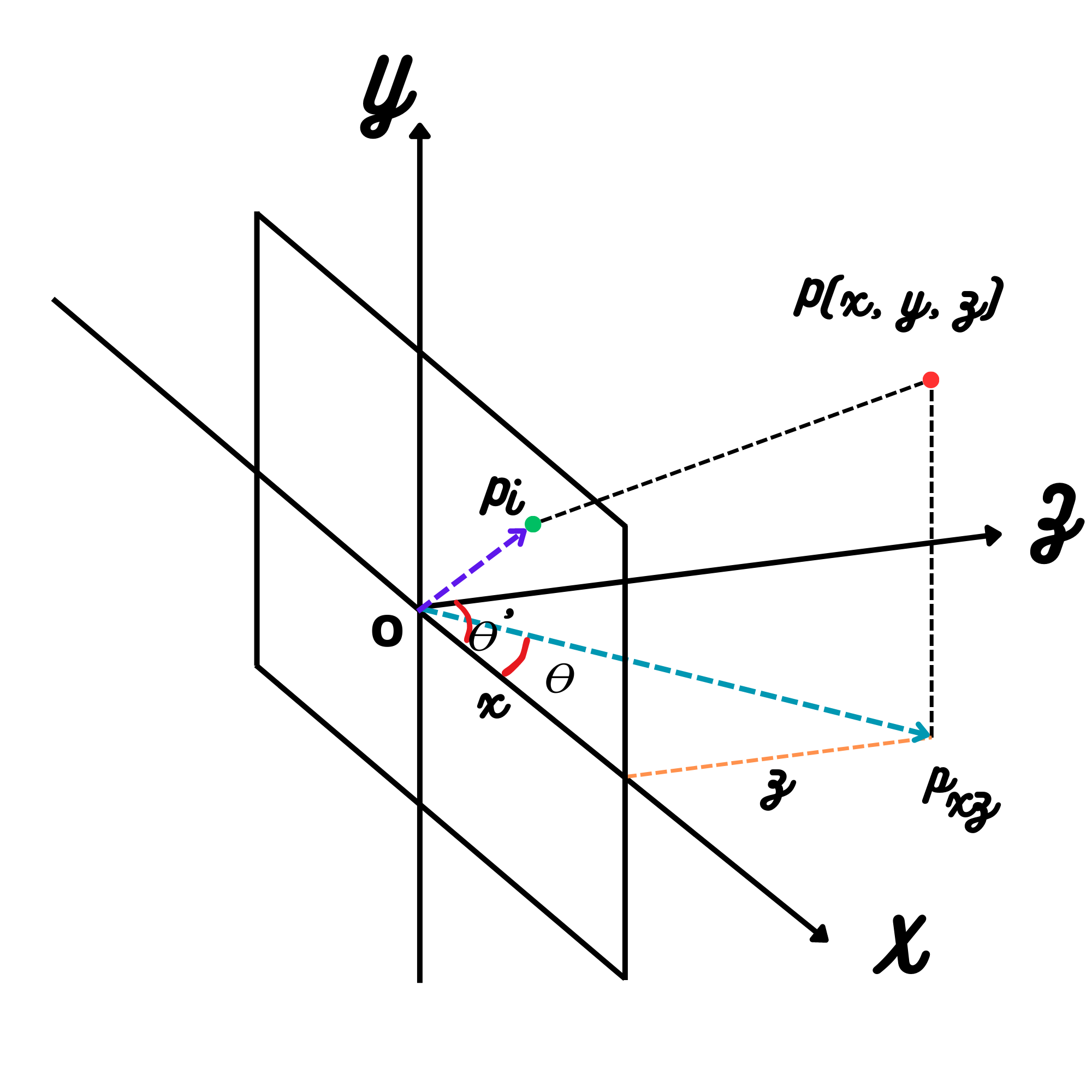}
    \caption{$p_i$ is the projection of the point $p$ into the image. The angle we want to compute is the angle formed by the object and the $(X)$ axis in the real world, not the axis of the image.}
    \label{fig:angles}
\end{figure}

Without the depth information and relying only on pixels coordinates, we would get an angle $\theta$'$=arctan(\frac{y_i}{x_i})$ where $(x_i, y_i)$ are the coordinates of the point $p$ on the image. $\theta$' is not the real angle of the direction of the object, but the angle formed by a projected centroid and the $(X)$ axis of the image as shown in figure \ref{fig:angles}.

\begin{figure}[!ht] 
    \centering
    \includegraphics[width=6cm]{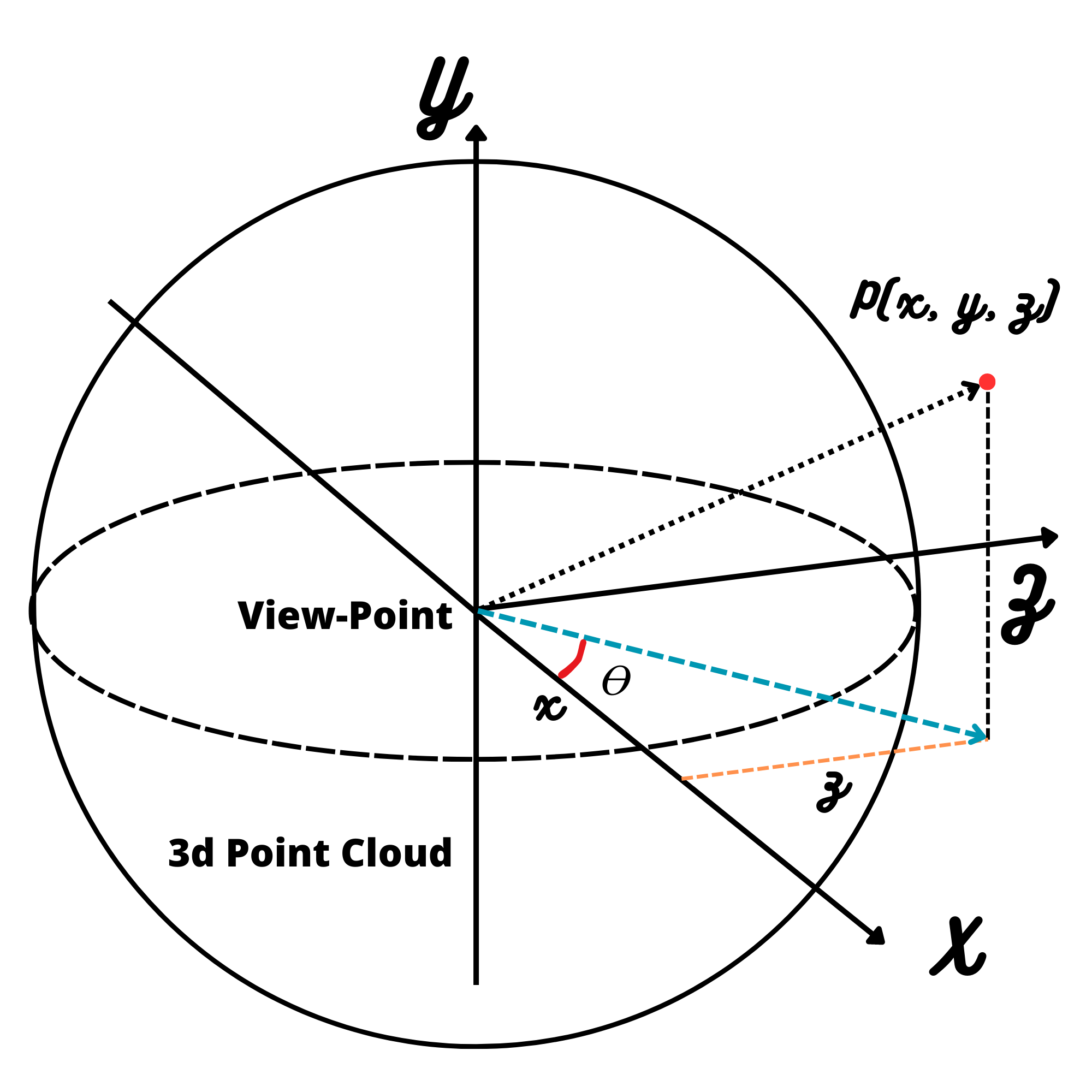}
    \caption{To compute the correct angle $\theta$ we built the point cloud, then we used the spherical coordinates to get the tangent of the angle formed by an object's centroid and the $(X)$ axis.}
    \label{fig:sphere}
\end{figure}

\begin{figure}[!ht] 
    \centering
    \includegraphics[width=7cm]{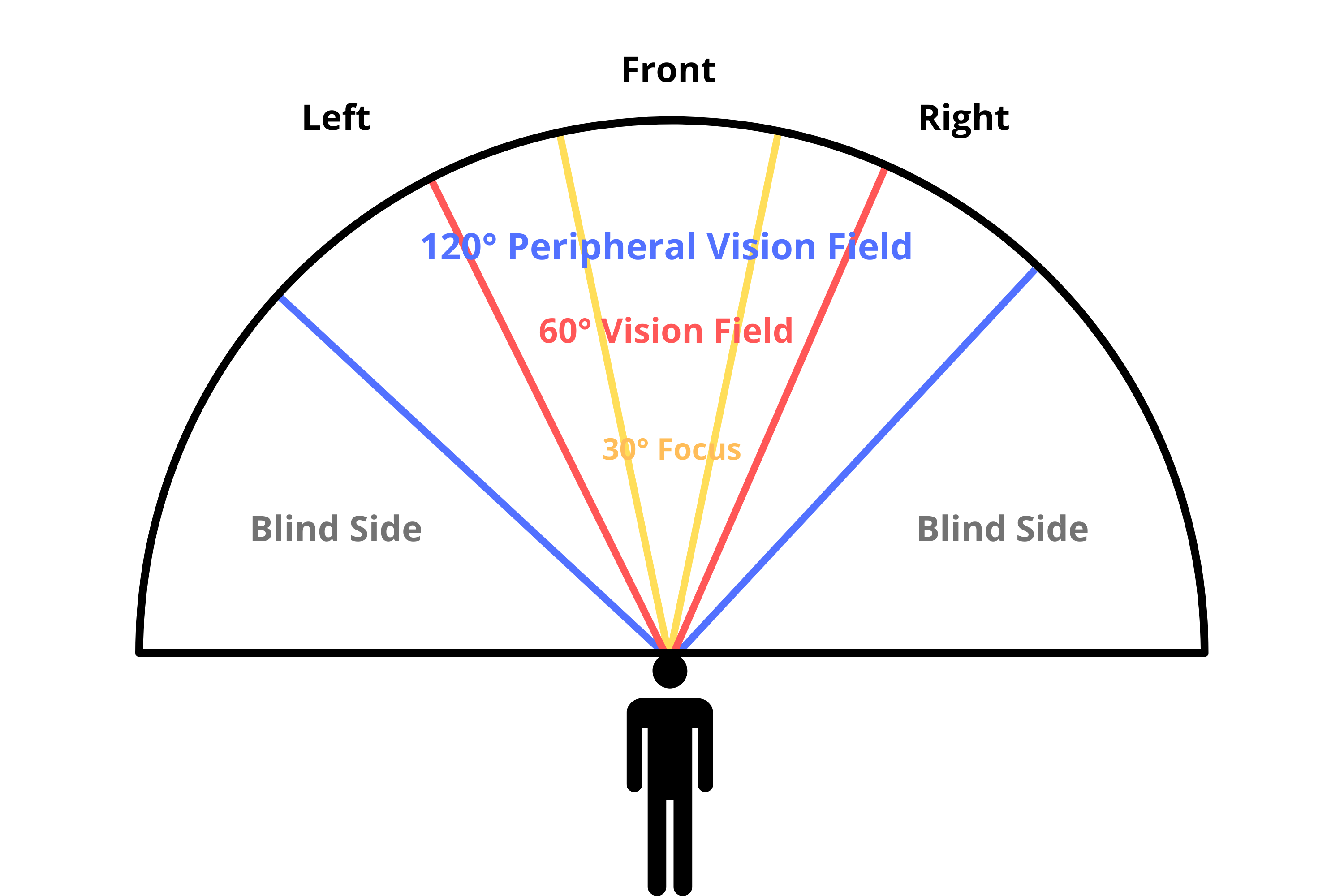}
    \caption{Fields of vision of humans, 30 degrees focus upfront, 45 degrees for left and 45 degrees for the right.}
    \label{fig:fields}
\end{figure}

In order to get the direction we followed the steps:
\begin{enumerate}
    \item Compute the centroid $(x, y, z)$ of each region, where $x$, $y$, and $z$ are the mean values of coordinates and depth values from the polygonal envelopes;
    \item Transform the values of the centroids to point-cloud coordinates using the built-in focal length parameter $f$ of the Microsoft Kinect v1, and the translation of the image plane to the 3D real-world plane $c_x$:
    $$
        x_w = (x - c_x) \times \frac{z}{f}
    $$
    \item Use the spherical coordinates system to compute the angle that the orthogonal projection of the region’s centroid into the $(XZ)$ plane forms with the $(X)$ axis:
    $$
        \theta=arctan(\frac{z}{x_w})
    $$
    the given result is the angle $\theta \in [-\frac{\pi}{2},\frac{\pi}{2}]$, as shown in figure \ref{fig:sphere};
    \item Use $\theta$  to determine to which field each region belongs, using the algorithm \ref{alg:directions} that is based on the human visual field from figure \ref{fig:fields}. 
\end{enumerate}

\begin{algorithm}[!t]
\caption{Directions from Angles}\label{alg:directions}
\begin{algorithmic}[1]
\Require angles, segment\_captioning\_result
\State $front [], right[], left[]$
\For{$region \in segment\_captioning\_result$}
    \State $\theta \gets region.angle$
    \State $\theta \gets degrees(\theta)$
    \If{$ 0 \leq \theta \leq 75$}
        \State $right.add(region.id)$
    \ElsIf{$0 \geq \theta \geq -75$}
        \State $left.add(region.id)$
    \Else
        \State $front.add(region.id)$
    \EndIf
\EndFor
\State \textbf{return} $front, right, left$
\end{algorithmic}
\end{algorithm}

Finally, we would have three sets of segments with captions for each field: \textit{Right, Left,} and \textit{Front} with 30 degrees of upfront focus.

\subsection{Determining Positional Relationships}
Determining the field to which each region or segment belongs already decreases the number of operations we need to process to get positional relationships between objects. It is trivial that objects on the right cannot overlap with objects on the left for example.

Furthermore, the depth maps that we have allow us to develop an algorithm that detects intersections between 3D bounding boxes without resorting to deep learning methods, making our solution less complex in terms of data processing and execution time.

Using the 3D bounding boxes, we create three adjacency matrices (left, right, front) where cells of overlapping objects are updated to contain the weight for each positional relationship: \textit{on the front of, behind, on, above, next to and under.} 

\begin{figure*}[!t]
\scalebox{0.8}{
\begin{minipage}{1.2\linewidth}
\centering
\subfloat{
    \subfloat{\includegraphics[height=3in, width=3in]{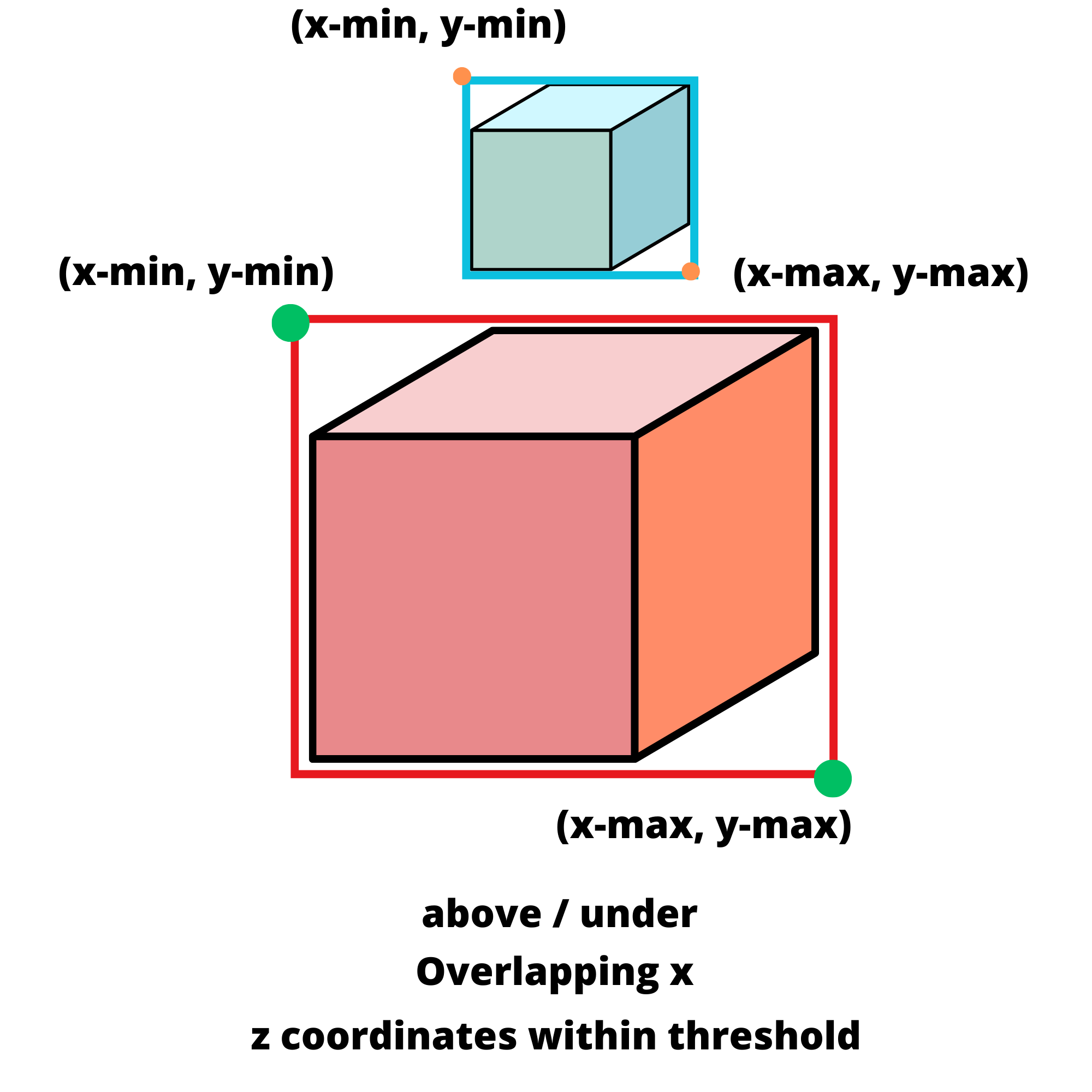}}
    \subfloat{\includegraphics[height=3in,width=3in]{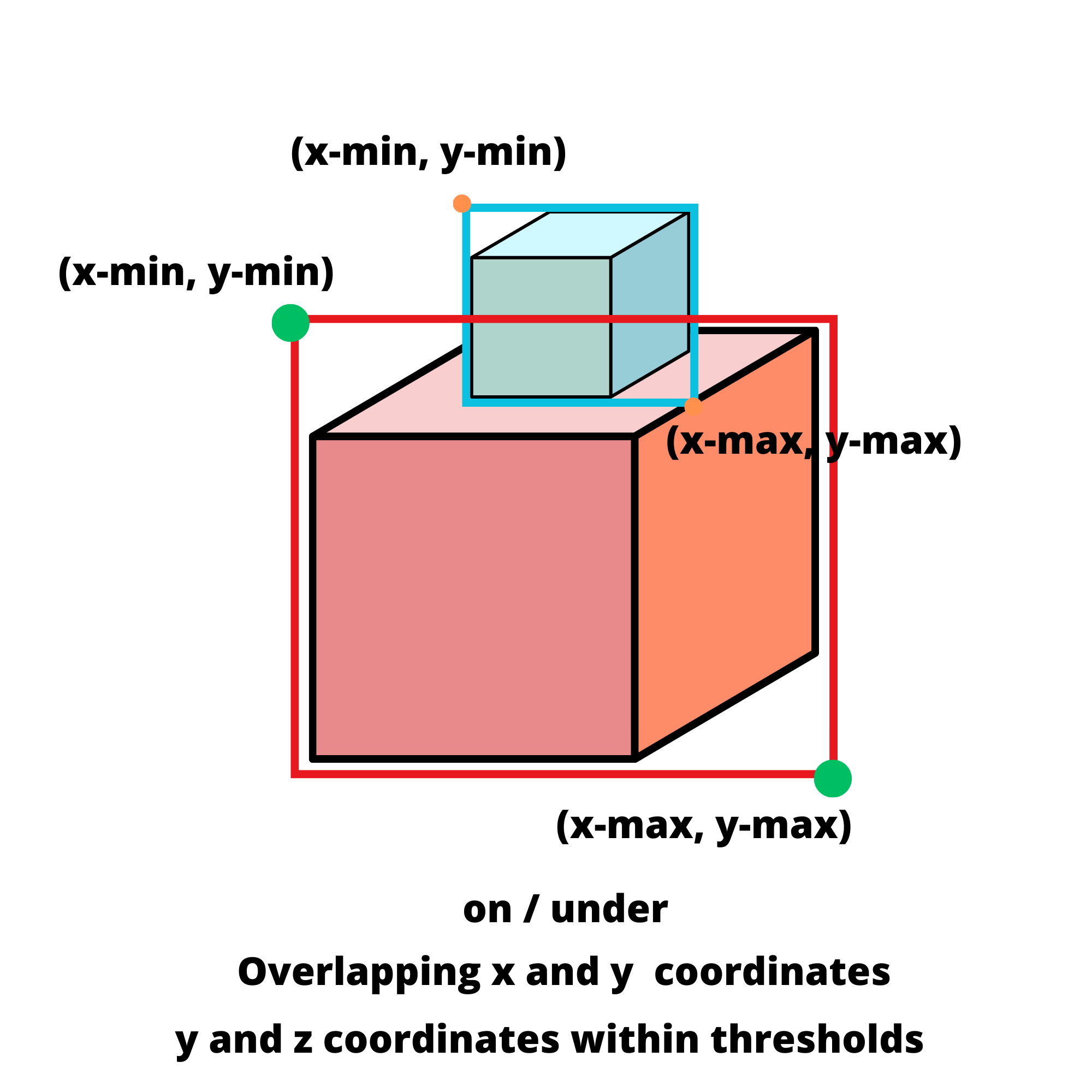}}
 
 }
\hfil
\subfloat{
    \subfloat{\includegraphics[height=3in, width=3in]{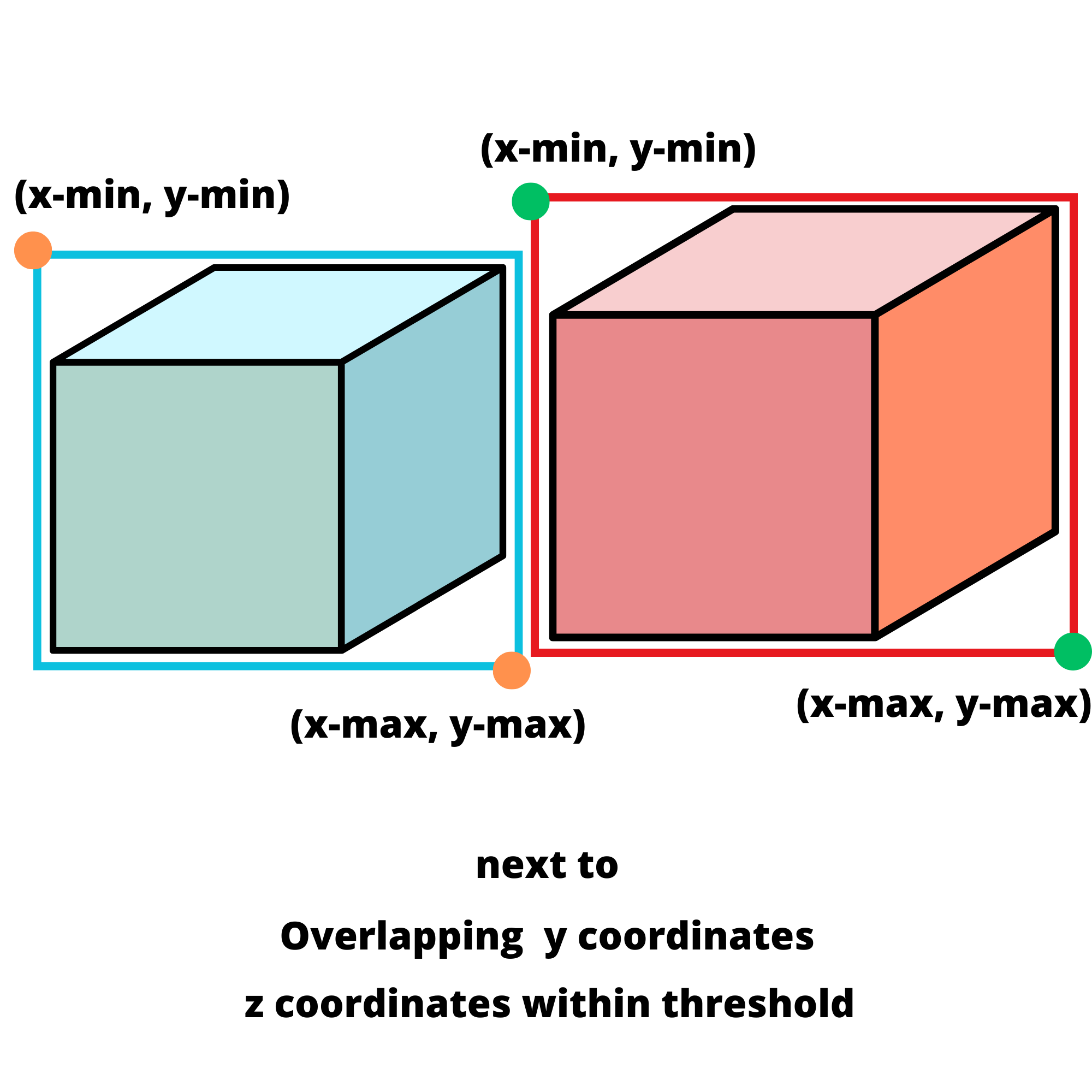}}
    \subfloat{\includegraphics[height=3in,width=3in]{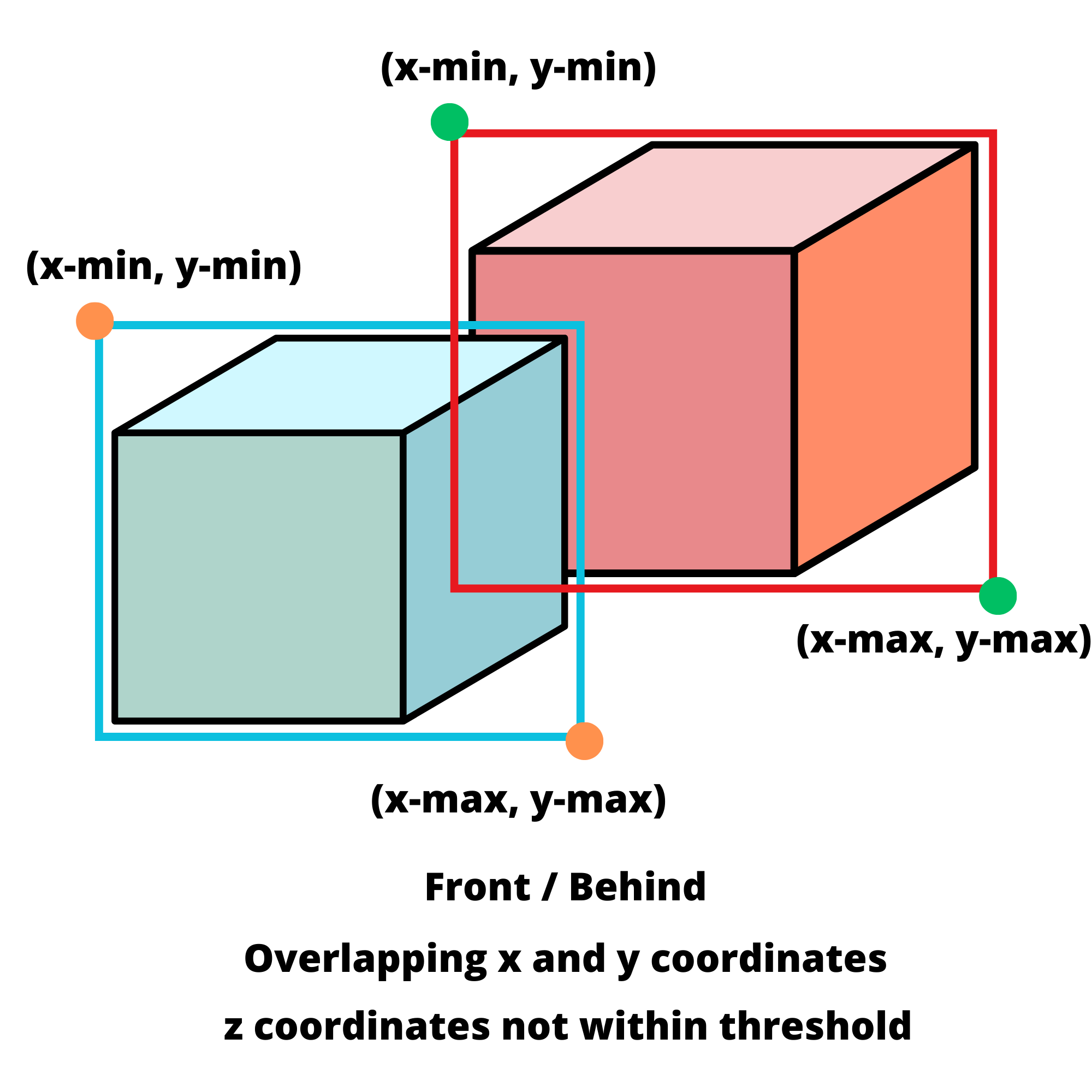}}
 }
\hfil
    \caption{The different positional relationships between objects.}
    \label{fig:cases}
\end{minipage}
}
\end{figure*}

The algorithm \ref{alg:relations} describes the process and the cases are summarised in figure \ref{fig:cases} which shows the basis on which the spatial relationships algorithm determines the position of a box regarding another.

\begin{algorithm}[!t]
\caption{Positional Relationships}\label{alg:relations}
\scalebox{0.8}{
\begin{minipage}{1.2\linewidth}
\begin{algorithmic}[1]
\Require $graph[][], regions\_in\_direction, nb\_regions, threshold\_Z, threshold\_Y$
\For{$i \in range(nb\_regions)$}
    \For{$ j \in range(i+1, nb\_regions)$}
            \State $ box1 \gets regions\_in\_direction[i]$
            \State $ box2 \gets regions\_in\_direction[j]$
            \State $x \gets Overlapping\_X(box1, box2)$
            \State $y \gets Overlapping\_Y(box1, box2)$
            \State $within\_z \gets |z_{mean_1} - z_{mean_2}| < threshold\_Z$
            \State $within\_y \gets |c_{y_1} - c_{y_2}| < threshold\_Y$
            \If{$x$ and $y$}
                \If{$within\_z$ and $within\_y$}
                    \IIf{$c_{y_1} > c_{y_2}$}  $graph[i, j] = on$
                    \ElseIIf $graph[i, j] = under$ 
                    \EndIIf
                \ElsIf{not $within\_z$}
                    \IIf{$z_{mean_1} > z_{mean_2}$} $graph[i, j] = behind$
                    \ElseIIf $graph[i, j] = front$
                    \EndIIf
                \Else 
                    \State $graph[i, j] = next\_to$
                \EndIf
            \ElsIf{$y$}
                \If{$within\_z$} $graph[i, j] = next\_to$
                \EndIf
            \Else 
                \If{$within\_z$} 
                    \IIf{$c_{y_1} > c_{y_2}$}  $graph[i, j] = above$
                    \ElseIIf $graph[i, j] = under$ 
                    \EndIIf
                \EndIf
            \EndIf
    \EndFor
\EndFor
\State \textbf{return} $graph$
\end{algorithmic}

\end{minipage}
}
\end{algorithm}

Lastly, we convert the matrix values into textual representations by employing the Euler Path Algorithm to analyse the adjacency matrix for each direction. These textual descriptions are then combined with the generated captions, resulting in a comprehensive scene description.

\section{Experiments and Results}
\label{sec:expr}
\subsection{Working Environment}
On a machine with AMD Ryzen 5700X 8 Core 3,4GHz CPU, 16 Go RAM, 2060 RTX Nvidia 8Go GPU, and Windows 11 as the operating system, training was performed on GPU with the latest versions of PyTorch Libraries (Python 3.9.7).

The PyTorch version of DenseCap provided in GitHub \cite{densecappytorch} served as a foundational framework for both our solutions.

\subsection{Dataset and Data Preprocessing}
We used our newly developed dataset of RGB-D Theatre scenes, the TS-RGBD dataset, that we collected using Microsoft Kinect v1 \cite{ourdataset}.  

\begin{figure}[!ht]
    \scalebox{0.8}{
    \begin{minipage}{1.2\linewidth}
    \centering
    \includegraphics[width=8cm]{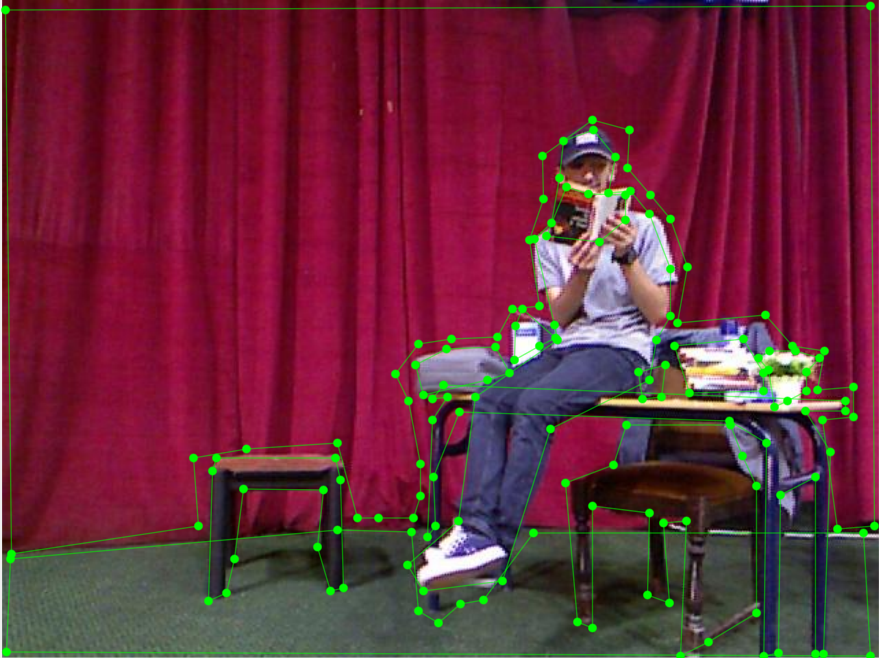}
    \includegraphics[width=8cm]{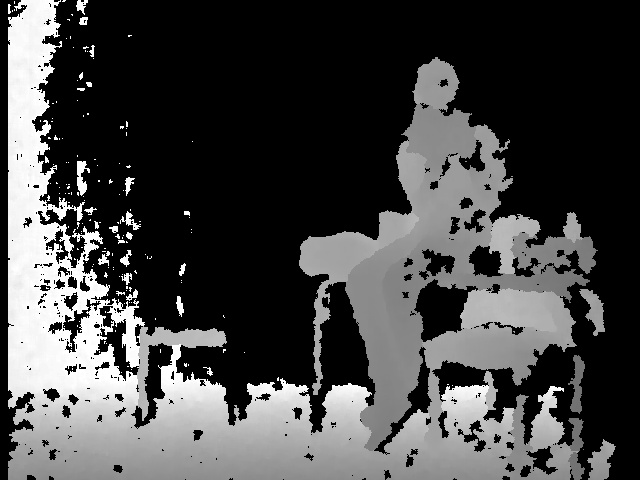}
    \includegraphics[width=14cm]{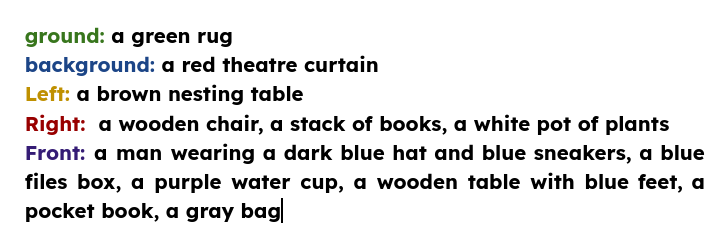}
    \caption{Example of collected and annotated data. Polygonal envelopes were defined by manually drawing points that contour a region of interest (green points).}
    \label{fig:annotated}
    \end{minipage}
    }
\end{figure}

We annotated our data manually using the “LabelMe” open-access framework \cite{labelme}. We had to manually draw polygonal envelopes for each region, and to replace labels with phrases. For each image, it generated JSON files that we preprocessed to build datasets with a dictionary for tokens, and bounding boxes for each polygonal envelope. Figure \ref{fig:annotated} shows a summary.

We trained the models on 120 images with a total of 1183 sentences, 109 different words. We split data into two sets, 50\% for training and 50\% for validation, with random data distribution.

The Microsoft Kinect v1 was used to capture depth values, resulting in depth maps that exhibit non-smooth characteristics. These maps may contain multiple 0 values and 4000 values (millimeters) that are considered noise, which can potentially affect the accuracy of mean, maximum, and minimum values derived from them.

To eliminate this problem, the mean filter was applied to depth maps to smooth such values.

\subsection{Panoptic Segmentation}
As for panoptic segmentation, numerous recent models are available within the open-access community. 

Unfortunately, we were unable to use the most recent real-time panoptic segmentation model proposed in \cite{yoso} due to compatibility issues with our working environment and the lack of support for our platform.

\begin{figure}[!ht]
    \centering
    \subfloat{
        \subfloat{\includegraphics[height=1.75in, width=3in]{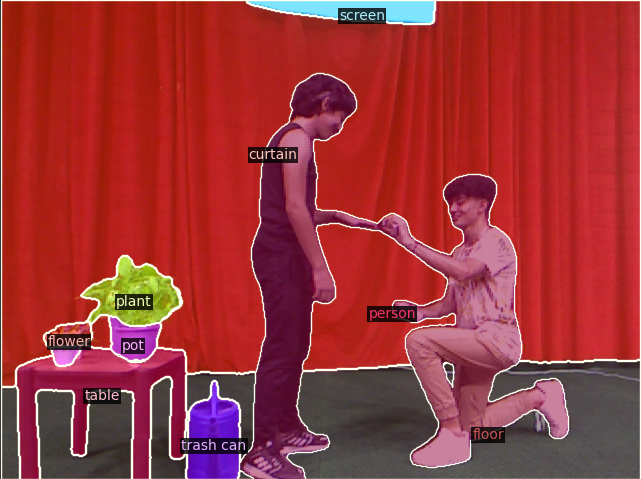}}
        \subfloat{\includegraphics[height=1.75in,width=3in]{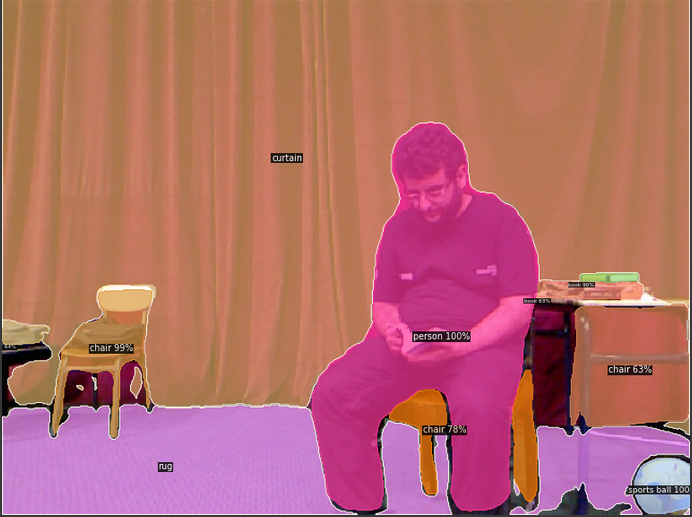}}
     }
    \hfil
    \caption{Examples of panoptic segmentation on theatre images using the OneFromer Model. }
    \label{fig:panoptic}
\end{figure}

We used the OneFormer \cite{oneformer}, a recent model that gave the best qualitative results for our dataset; figure \ref{fig:panoptic}. It relays mainly on transformer architecture without resorting to RPN models.

When testing the whole framework, we don't use ground truth boxes, so for a segmented image, we would get the bounding boxes of each segment by creating for each segment a binary image where segment pixels are equal to one.

Since segments are inferred by a model, there could be some other pixels outside of a segment with the same $id$, so we contoured all white regions and took the largest one among them. For the largest contoured area, we extract the bounding box.

\subsection{Region Captioning Results}
We conducted experiments on our proposed architectures with different parameters and hyperparameters.

In the following section, we will showcase and analyze the various results we have achieved in our study.

\subsubsection{Parameters, Hyper-Parameters, Loss Function, and Evaluation Metrics}
The training was conducted on different instances of our dataset. First, we considered sentences no longer than 15 words, then sentences with a maximum length of 10 words. The latter gave better results. For each sentence length, we considered tokens that appeared at least 2 times and those with only one appearance. The first gave better results.

As for hyper-parameters, we used Adam Optimizer, torch Grad Scaler, and learning rate.

The cross-entropy loss function is chosen for the captioning module and the softmax function for the output prediction.

Evaluation metrics for the language are BLEU, ROUGE, and CIDEr.

\subsubsection{Training from Scratch}
Training on our dataset led to an important overfitting due to the lack of data. The LSTM model could generate accurate captions but would also add unnecessary tokens to each phrase due to overfitting. 

Although the loss curves of both models (head 1 and head 2) were decreased, the evaluation curves stabilised after a number of epochs. The table \ref{tab:captions_results} shows results obtained from both heads on our images after 100 epochs.

The quantitative results showed that the features extractor plays an important role in the accuracy of captions. When using a feature extractor on the region only, it slows the process and decreases accuracy. In addition, the FPN feature extractor gives better results than the VGG-16.

Due to the evident problem of overfitting, we decided to explore transfer learning as a potential solution.

\subsubsection{Transfer Learning}
After considering the advantages of the second head,  we opted for a pre-trained Feature Pyramid Network (FPN) with a pre-trained ResNet-50 backbone. We selectively blocked back-propagation to ensure the desired updates, allowing only the intermediary layer and the LSTM to be updated.

Results are shown in table \ref{tab:captions_results}.

\begin{figure}[!ht]
    \centering
    \includegraphics[width=8cm]{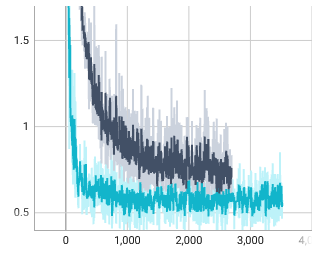}
    \caption{Loss curves from training our model: dark blue is the loss curve of training from scratch, light blue is the loss curve of the transfer learning. }
    \label{fig:transfer}
\end{figure}

\begin{figure}[!ht]
    \centering
    \includegraphics[width=8cm]{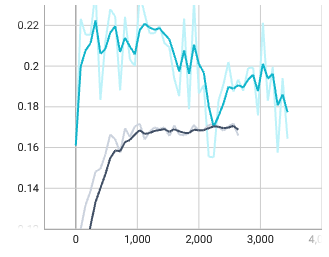}
    \caption{BLEU metric curves form training our model: dark blue is the BLEU results curve of training from scratch, and light blue is for transfer learning. }
    \label{fig:transfer_bleu}
\end{figure}

Figure \ref{fig:transfer} shows the obtained loss curve and figure \ref{fig:transfer_bleu} the accuracy using the BLEU metric.

Although the accuracy of the results improved noticeably, the overall shape of the learning curve remained unchanged. Both training approaches yielded stabilised results, indicating that the LSTM is stuck in a repetitive loop and failing to converge. The oscillating shape of the evaluation curve confirms that.

These results led to our training phase's final step, Fine Tuning.

\subsubsection{Fine Tuning}
Before developing our solution, we had already trained DenseCap on our data. So we retrieved the weights obtained from the training, and we fed our model with FPN, Resnet-50 Backbone, Intermediary Layer, and LSTM weights as starting points.

After only 10 epochs, the LSTM converged successfully and the accuracy reached the results shown in table \ref{tab:captions_results}.

\begin{table*}[!ht] 
\caption{Results on different metrics, for our models, the higher the better.}
\label{tab:captions_results}
\begin{tabular*}{\linewidth}{l @{\extracolsep{\fill}}c c c c c c}
\toprule
Metrics & BLEU@1 & BLEU@2 & BLEU@3 & BLEU@4 & ROUGE & CIDEr \\
\midrule
\multicolumn{7}{c}{Training from Scratch}\\
\midrule
Head-1 & 0.03 & 0.01 & 0.012 & 0.001 & 0.09 & 0.015 \\
Head-2 & 0.17 & 0.11 & 0.09 & 0.07 & 0.22 & 0.23 \\ 
\midrule
\multicolumn{7}{c}{transfer Learning}\\
\midrule
Head-2 & 0.22 & 0.14 & 0.11 & 0.08 & 0.56 & 0.25 \\
\midrule
\multicolumn{7}{c}{Fine Tuning}\\
\midrule
\textbf{Head-2} & \textbf{0.39} & \textbf{0.34} & \textbf{0.31} & \textbf{0.29} & \textbf{3.25} & \textbf{0.47} \\
\bottomrule
\end{tabular*}
\end{table*}

\begin{figure}[!ht]
    \centering
    \includegraphics[width=8cm]{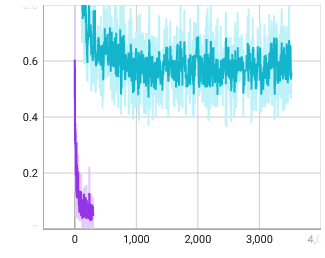}
    \caption{Loss curves from training our model: light blue is the loss curve of the transfer learning, purple is the loss curve of the fine-tuning after only 10 epochs. }
    \label{fig:finetuning}
\end{figure}

\begin{figure}[!ht]
    \centering
    \includegraphics[width=8cm]{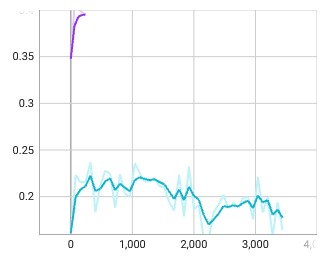}
    \caption{BLEU metric curves from training our model: light blue is the loss curve of the transfer learning, purple is the loss curve of the fine-tuning after only 10 epochs. }
    \label{fig:fine_bleu}
\end{figure}

\begin{figure}[!ht]
    \centering
    \includegraphics[width=8cm]{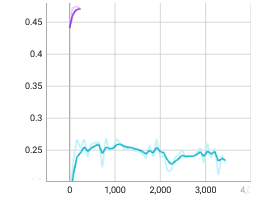}
    \caption{ROUGE metric curves form training our model: light blue is the loss curve of the transfer learning, purple is the loss curve of the fine-tuning after only 10 epochs. }
    \label{fig:fine_rouge}
\end{figure}

\begin{figure}[!ht]
    \centering
    \includegraphics[width=8cm]{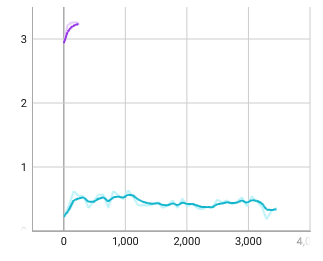}
    \caption{CIDEr metric curves form training our model: light blue is the loss curve of the transfer learning, purple is the loss curve of the fine-tuning after only 10 epochs. }
    \label{fig:fine_cider}
\end{figure}

Figure \ref{fig:finetuning} shows how 10 epochs were enough to get a smaller loss value with a better shape of the loss curve. Figures \ref{fig:fine_bleu}, \ref{fig:fine_cider} and \ref{fig:fine_rouge} show how the accuracy using the BLEU, ROUGE, and CIDEr metrics when fine-tuning gave largely better results.

\begin{figure}[!ht]
    \centering
    \includegraphics[width=12cm]{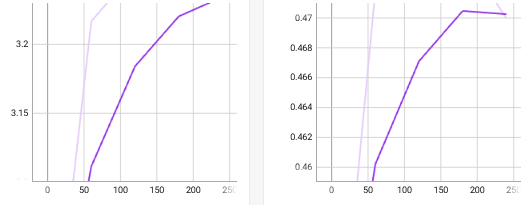}
    \caption{Curves of ROUGE and CIDEr results after only 10 epochs for fine-tuning. }
    \label{fig:fine_cider_rouge}
\end{figure}

Figure \ref{fig:fine_cider_rouge} shows how the accuracy is increasing while keeping a smooth and steady shape without oscillations.

\subsection{Comparative Study}
Table \ref{tab:exec_densecap_ours} shows the average test run time for the DenseCap model and our Segment Captioning model on our environment.

\begin{table}[!ht]
    \renewcommand{\theadfont}{\normalsize}
    \settowidth\rotheadsize{\theadfont boredom}
    \setlength{\tabcolsep}{0pt}
    \caption{Execution time comparison between DenseCap and Our Model}
    \label{tab:exec_densecap_ours}
    \begin{tabular*}{\linewidth}{@{\extracolsep{\fill}}l  c c cc@{}}
        \toprule
        Model & RPN & CNN & LSTM & Total per Image \\
        \midrule
        DenseCap & 36ms & 55 ms &  4ms & ~95ms \\
        \textbf{Ours} & - & 48ms & 2ms & \textbf{~50ms} \\
        \bottomrule
    \end{tabular*}
\end{table}

The reduction in processing time can be attributed to the removal of the Localization Layer, which incorporates an RPN (Region Proposal Network). Instead of spending time to find boxes with high scores, our model takes the coordinates of bounding boxes as input. 

Additionally, our Features Extractor has become marginally faster as a result of reducing the number of regions that require computing the RoI Align.

To study the performance of our model and DenseCap model when used as captioners for our whole framework, we had to remove boxes proposed by DenseCap and keep only the regions that align with segments.

In order to remove unwanted boxes from the results of DenseCap and retain only those that align with the regions provided by panoptic segmentation, we calculate the intersection over union scores for each region and select the best matching boxes based on these scores.

The table \ref{tab:exec_densecap_ours_final} shows the execution time for the general region captioning.

\begin{table}[!ht]
    \renewcommand{\theadfont}{\normalsize}
    \settowidth\rotheadsize{\theadfont boredom}
    \setlength{\tabcolsep}{0pt}
    \caption{Execution time comparison between DenseCap and Our Model when eliminating unnecessary regions.}
    \label{tab:exec_densecap_ours_final}
    \begin{tabular*}{\linewidth}{@{\extracolsep{\fill}}l  c cc@{}}
        \toprule
        Model & Inference & Box Align & Total per Image \\
        \midrule
        DenseCap & ~95ms & 135 ms & 225ms  \\
        \textbf{Ours} & \textbf{~50ms} & - & \textbf{~50ms}  \\
        \bottomrule
    \end{tabular*}
\end{table}

\textbf{\textit{In conclusion, our model is better when it comes to Region/Segment Captioning as it decreases the execution time $T$.
}}

$$
\boxed{T(Seg) + T(Ours) <<  T(Seg) + T(DenseCap) + T(Box Align)}
$$

\subsection{Final Captions}
By running the previously detailed algorithms \ref{alg:directions} and \ref{alg:relations}, we would assign to each region a direction and relationships with overlapping objects. Objects on the ground and in the background are kept apart. The object with the maximum $c_y$ value ($y$ coordinate of its centroid) is the object on the ground. The object with the maximum depth value $z_{mean}$ (the mean value of its overall depth) is the object in the background. 

Thresholds from the algorithm \ref{alg:relations} were chosen by trial and error in order to determine if an object could be considered close and in intersection with another one.

For our annotated data with 1183 regions, the table \ref{tab:final_res} summarises the achieved results.

\begin{table}[!ht]
    \renewcommand{\theadfont}{\normalsize}
    \settowidth\rotheadsize{\theadfont boredom}
    \setlength{\tabcolsep}{0pt}
    \caption{Results of our final framework in terms of execution time and accuracy.}
    \label{tab:final_res}
    \begin{tabular*}{\linewidth}{@{\extracolsep{\fill}}*6c@{}}
        \toprule
        \multicolumn{2}{c}{Captioning} & \multicolumn{2}{c}{Directions} & \multicolumn{2}{c}{Positions} \\
        \midrule
        Time & BLEU & Time & Acc. & Time & Acc. \\
        \midrule
         50ms & 0.332 & 1e-5ms & 99\% & 1e-4ms & 90\% \\
        \bottomrule
    \end{tabular*}
\end{table}

\begin{figure*}[!t]
\scalebox{0.7}{
\begin{minipage}{1.35\linewidth}
\centering
\includegraphics[width=8cm]{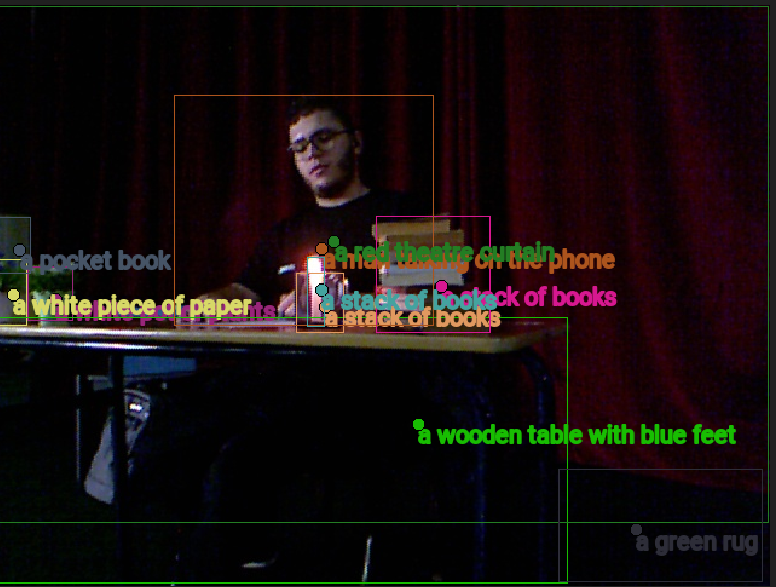}
\includegraphics[width=8cm]{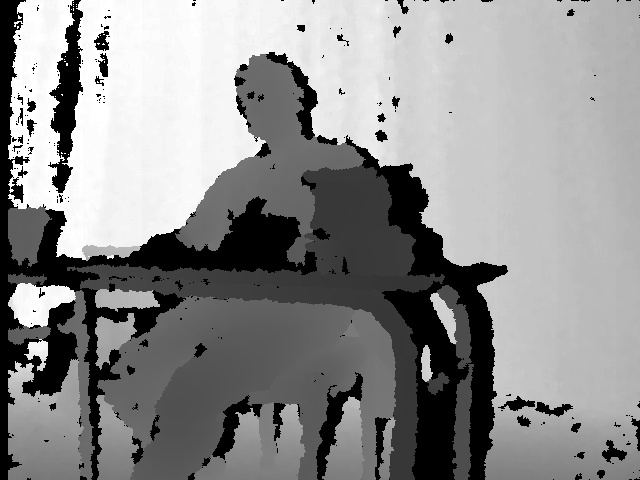}
 \includegraphics[width=8cm]{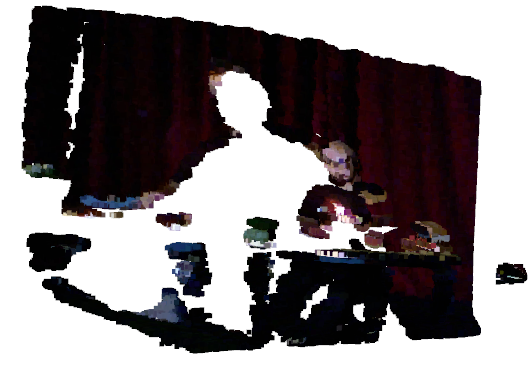}
\includegraphics[width=14cm]{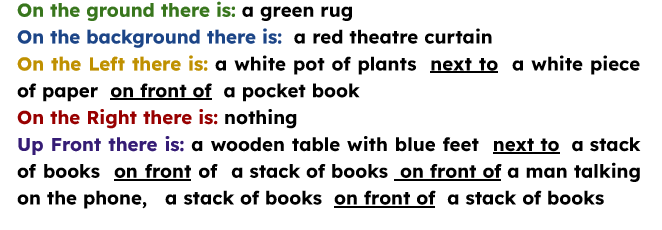}
 \caption{First Example from our Theatre Scenes RGBD dataset.}
    \label{fig:results1}
    \end{minipage}
    }
\end{figure*}

\begin{figure*}[!t]
\scalebox{0.7}{
\begin{minipage}{1.35\linewidth}
\centering
\includegraphics[width=8cm]{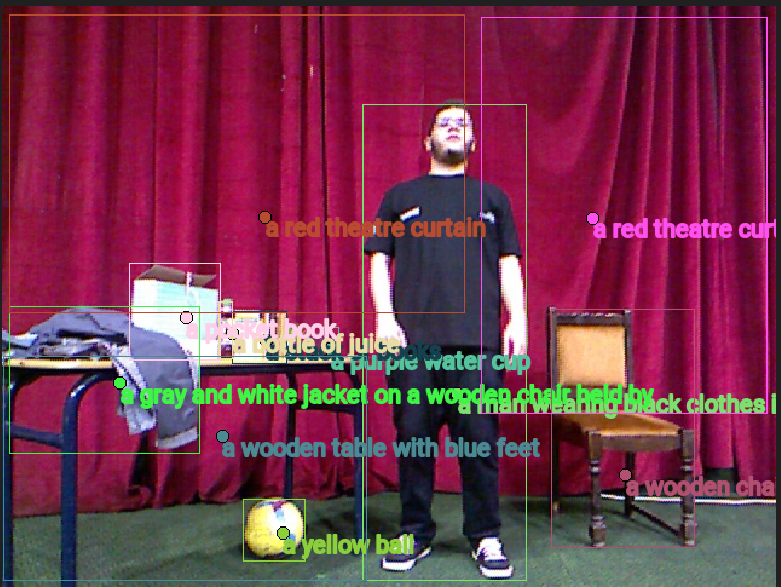}
\includegraphics[width=8cm]{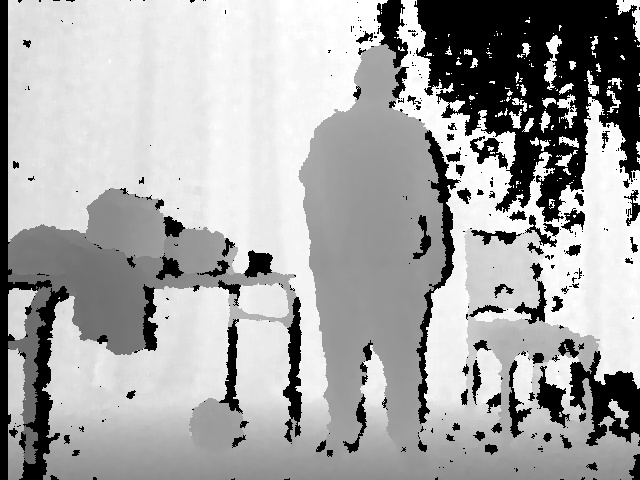}
    \includegraphics[width=8cm]{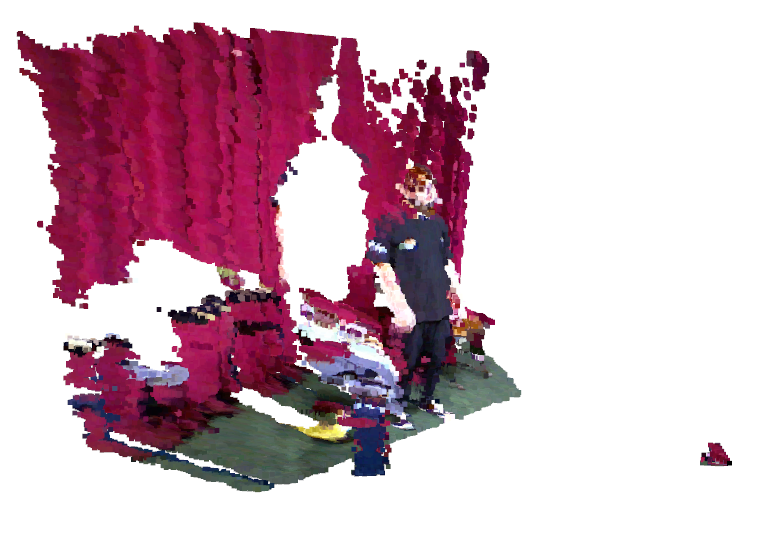}
\includegraphics[width=14cm]{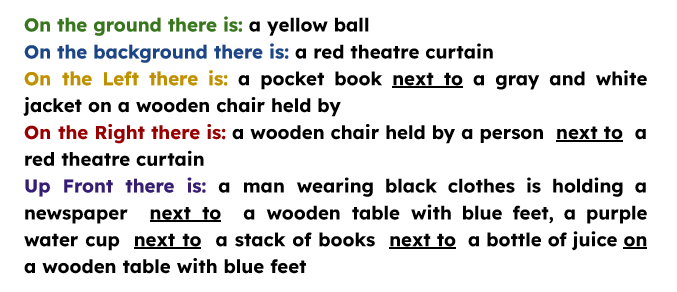}
 \caption{Second Example from our Theatre Scenes RGBD dataset.}
    \label{fig:results2}
    \end{minipage}
    }
\end{figure*}

\begin{figure*}[!t]
\scalebox{0.7}{
\begin{minipage}{1.35\linewidth}
\centering
\includegraphics[width=8cm]{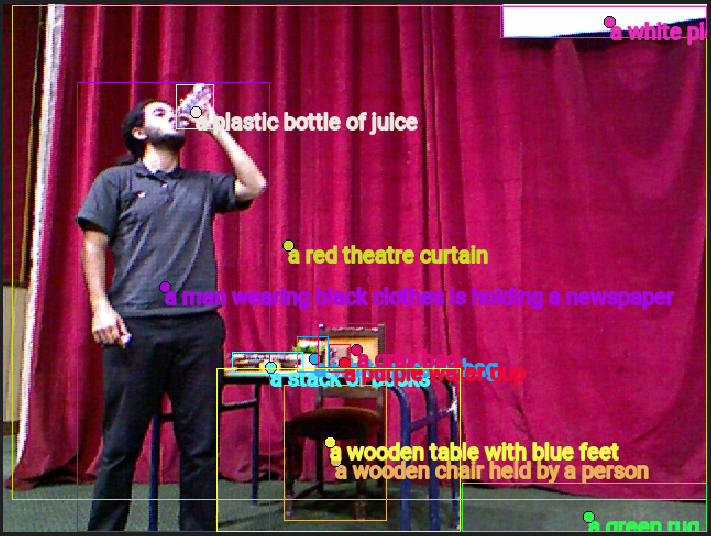}
  \includegraphics[width=8cm]{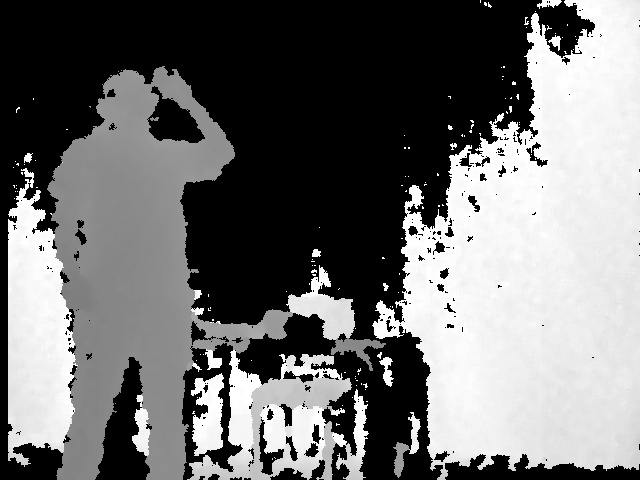}
  \includegraphics[width=8cm]{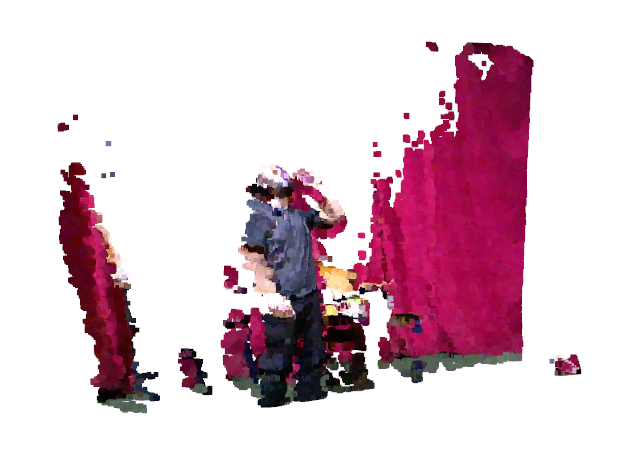}
\includegraphics[width=14cm]{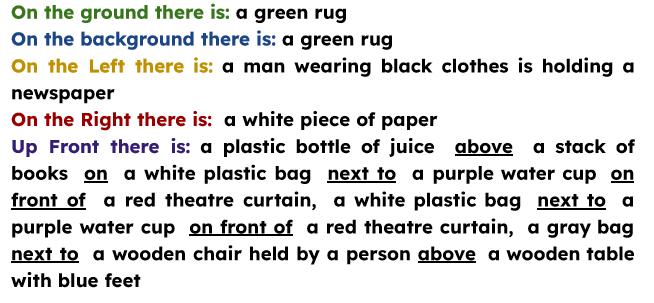}
 \caption{Third Example from our Theatre Scenes RGBD dataset.}
    \label{fig:results3}
    \end{minipage}
    }
\end{figure*}

Qualitative results are shown in the figures \ref{fig:results1}, \ref{fig:results2} and \ref{fig:results3}.

Generated captions are satisfactory regarding the accuracy achieved on different metrics. Even in images with less lighting showing the effectiveness of the model. 
Egocentric descriptions are correct because they are based on the computation of a real-world angle using depth values.

Positional relationships are accurate compared to ground truth positions, confirming the effectiveness of the developed algorithm.

Even if some directions or positional relationships would seem false when looking only at the RGB image, they are correct compared to ground truth. This is due to perspectives that change after capturing a 3D world scene into a 2D image, hence the need of the depth maps.

% \hfil
% \subfloat{
%     \subfloat{\includegraphics[height=1in,width=2in]{figures/191.png}}
%     \subfloat{\includegraphics[height=1in,width=2in]{figures/12604.jpeg}}
%     \subfloat{\includegraphics[height=1in,width=2in]{figures/point_cloud_191.png}}
%  }
% \hfil
% \subfloat{
%     \subfloat{\includegraphics[height=1in,width=4in]{figures/captions191.png}}
% }
% \hfil
% \subfloat{
%     \subfloat{\includegraphics[height=1in,width=1.5in]{figures/989.png}}
%     \subfloat{\includegraphics[height=1in,width=1.5in]{figures/46519.jpeg}}
%     \subfloat{\includegraphics[height=1in,width=1.5in]{figures/point_cloud_989.png}}
%     \subfloat{\includegraphics[height=1in,width=1.5in]{figures/captions989.png}}
    
% }
% \hfil
% \subfloat{
%     \subfloat{\includegraphics[height=1in,width=1.5in]{figures/1484.png}}
%     \subfloat{\includegraphics[height=1in,width=1.5in]{figures/98571.jpeg}}
%     \subfloat{\includegraphics[height=1in,width=1.5in]{figures/point_cloud_1484.png}}
%     \subfloat{\includegraphics[height=1in,width=1.5in]{figures/captions1484.png}}
% }
\hfil

\subsubsection{Discussion}
From qualitative, quantitative results and training curves:
\begin{itemize}
    \item The framework answered the problem of egocentric theatre scene description;
    \item Due to the small size of our dataset, the overfitting is still present and it is noticed when changing slightly the coordinates of bounding boxes, hence including more or fewer pixels for a region to be described, the LSTM captioner performance drops;
    \item From results given by Head 1 and Head 2, we deduce that information from pixels surrounding a bounding box is important for more contextual meaning since Head 1 eliminates pixels outside of the polygonal envelope gave worse results; 
    \item The directions algorithm performance is satisfying compared to ground truth;
    \item The positional relationships algorithm's performance was highly satisfactory, given the accuracy and the speed of the execution;
    \item The execution time of the whole framework is satisfactory.
\end{itemize}

\section{Limitations}
\label{sec:limitations}
In this paper, we presented an egocentric segment captioning framework that we tested on theatre images to offer users with visual disabilities an extension of the environments they can be in. 

Here are the flaws of our proposed solution:
\begin{enumerate}
    \item Due to the overfitting of the captioning model, this solution cannot be applied to new images;
    \item Due to the overfitting of the captioning model, if an input bounding box of a region is less accurate than the ground truth bounding box, or shifted, then the accuracy of the model drops;
    \item Our solution depends on panoptic segmentation, if the segmentation model gives bad results it would mislead the captioning model;
    \item Our solution depends on panoptic segmentation, if the used model is slow it would slow up the execution;
    \item The directions algorithm depends on the camera parameters when computing the point cloud;
    \item The positional relationships algorithm depends on two threshold values that determine whether an object is close enough to another to be in a spatial relationship.
\end{enumerate}

\section{Future Perspectives}
\label{sec:pers}
Our model has shown promising results on the small validation set, indicating its effectiveness. This encourages us to continue with data annotation and consider collecting additional data with more diverse and sophisticated materials, that involve real actors and disguises, which can further enhance the performance of our model. 

The accuracy of generated phrases and their correct semantics motivates us to refine our annotations for future work, incorporating more theatre-specific language such as using terms like "prince" instead of "man" and including female characters as "princesses" for example, along with other theatrical elements and settings. 

We also plan on retraining a fast segmentation model to keep up with the real-time execution and enriching captions with textual descriptions of actions by applying the methods proposed in \cite{benhamida} since they rely on depth as well.

Furthermore, we aim to extend our work to video processing, as our framework has demonstrated good execution time, achieving 24 frames per second.

Finally, the solution will be presented to actual users with visual disabilities for a better evaluation.

\section{Conclusion}
\label{sec:conclu}
In this article, we have provided a comprehensive review of the recent advancements in AI technologies for visual scene understanding. Through an extensive analysis of the literature, we have examined the state-of-the-art techniques and methodologies employed in various domains, including image captioning, image segmentation, and scene understanding.

Furthermore, we have discussed the challenges and limitations that still exist in visual scene understanding. Despite significant advancements, issues such as handling occlusions, robustness to diverse environmental conditions, and generalization to unseen scenarios remain areas of active research.

In our study, we addressed the challenge of developing a framework that answers the needs of blind and visually impaired individuals by providing textual descriptions of regions of interest within an image. Our objective was to generate captions that not only describe the visual attributes of these regions but also incorporate their relative positions with respect to the user and establish their positional relationships with one another.

We applied our solution to our new TS-RGBD dataset, a dataset of RGB-D images of theatre scenes, which is considered a novel field of application for image captioning.

Our solution proved to be more effective compared to other image captioning models, in terms of the number of captions per image and the execution time.
The positional relationships between objects and the egocentric descriptions were successfully processed by fast and simple algorithms, independent of machine learning methods. Those algorithms used the depth information from the depth maps to improve the accuracy of the descriptions.

For future work, we will improve the quality of image ground truth captions by making them more theatre-specific, and augment our TS-RGBD dataset by collecting and annotating additional data.
We will improve positional relationships extraction and we will use more sophisticated sensors to expand the field of capture and bring our project to actual theatre plays.

Our solution will be presented to blind and visually impaired users for a better evaluation of the generated captions.

%% The Appendices part is started with the command \appendix;
%% appendix sections are then done as normal sections

% \input{sections/appendix}

%% If you have bibdatabase file and want bibtex to generate the
%% bibitems, please use
%%
%%  \bibliographystyle{elsarticle-harv} 
%%  \bibliography{<your bibdatabase>}

%% else use the following coding to input the bibitems directly in the
%% TeX file.

%% If you have bibdatabase file and want bibtex to generate the
%% bibitems, please use
%%
%%  \bibliographystyle{elsarticle-harv} 
%%  \bibliography{<your bibdatabase>}

%% else use the following coding to input the bibitems directly in the
%% TeX file.

\end{document}